\documentclass[letterpaper]{article} %
\usepackage{aaai2026}  %
\usepackage{times}  %
\usepackage{helvet}  %
\usepackage{courier}  %
\usepackage[hyphens]{url}  %
\usepackage{graphicx} %
\usepackage{natbib}  %
\usepackage{caption} %
\usepackage{subcaption} %
\usepackage[ruled]{algorithm2e}
\usepackage[most]{tcolorbox}

\usepackage[shortcuts]{extdash}  %
\usepackage{comment}

\newcommand \planqd {PLAN\=/QD}

\usepackage[dvipsnames]{xcolor}

\definecolor{maroon}{RGB}{128,0,0}    %
\usepackage{soul}
\newcommand{\hidenotes}{} %

\newcommand{\vbnote}[1]{{\xxnote{VB}{blue}{#1}}}

\newcommand{\xxnote}[3]{}
\ifx\hidenotes\undefined
  \renewcommand{\xxnote}[3]{\color{#2}{(#1: #3)}}
\fi

\newcommand{\sref}[1]{Sec.~\ref{#1}}
\newcommand{\apref}[1]{App.~\ref{#1}}
\newcommand{\fref}[1]{Fig.~\ref{#1}}
\newcommand{\tref}[1]{Table~\ref{#1}}
\newcommand{\aref}[1]{Algorithm~\ref{#1}}

\usepackage{booktabs}
\usepackage{multirow}
\usepackage{array}
\usepackage{caption} 
\newcolumntype{L}[1]
  {>{\raggedright\let\newline\\\arraybackslash\hspace{0pt}}m{#1}}
\newcolumntype{C}[1]
  {>{\centering\let\newline\\\arraybackslash\hspace{0pt}}m{#1}}
\newcolumntype{R}[1]
  {>{\raggedleft\let\newline\\\arraybackslash\hspace{0pt}}m{#1}}

\usepackage{amsmath,amsfonts,bm}

\def\eqref#1{equation~\ref{#1}}

\def\1{\bm{1}}

\def\vm{{\bm{m}}}

\def\vx{{\bm{x}}}

\def\vz{{\bm{z}}}

\DeclareMathAlphabet{\mathsfit}{\encodingdefault}{\sfdefault}{m}{sl}
\SetMathAlphabet{\mathsfit}{bold}{\encodingdefault}{\sfdefault}{bx}{n}

\def\gA{{\mathcal{A}}}

\def\gP{{\mathcal{P}}}

\def\gR{{\mathcal{R}}}
\def\gS{{\mathcal{S}}}
\def\gT{{\mathcal{T}}}

\def\gX{{\mathcal{X}}}

\def\gZ{{\mathcal{Z}}}

\def\sR{{\mathbb{R}}}

\title{Algorithmic Prompt Generation for Diverse Human-like Teaming and Communication with Large Language Models}

\author{
    Siddharth Srikanth\textsuperscript{\rm 1},
    Varun Bhatt\textsuperscript{\rm 1},
    Boshen Zhang\textsuperscript{\rm 1},
    Werner Hager$^2$,
    Charles Michael Lewis$^2$,
    Katia P. Sycara$^3$,
    Aaquib Tabrez\textsuperscript{\rm 4},
    Stefanos Nikolaidis$^1$ 
}
\affiliations {
    \textsuperscript{\rm 1}Thomas Lord Department of Computer Science, University of Southern California\\
    \textsuperscript{\rm 2}School of Computing and Information, University of Pittsburgh\\
    \textsuperscript{\rm 3}Robotics Institute, Carnegie Mellon University\\
    \textsuperscript{\rm 4}Sibley School of Mechanical and Aerospace Engineering, Cornell University\\
    \texttt{\{ssrikant,vsbhatt,boshenzh,nikolaid\}@usc.edu},
    \texttt{aaquibtabrez@cornell.edu}, \\
    \texttt{WWH10@pitt.edu},
    \texttt{ml@sis.pitt.edu}, \texttt{sycara@andrew.cmu.edu}
}

\nocopyright
\begin{document}

\maketitle

\begin{abstract}

Understanding how humans collaborate and communicate in teams is essential for improving human-agent teaming and AI-assisted decision-making. 
However, relying solely on data from large-scale user studies is impractical due to logistical, ethical, and practical constraints, necessitating synthetic models of multiple diverse human behaviors. 
Recently, agents powered by Large Language Models (LLMs) have been shown to emulate human-like behavior in social settings.
But, obtaining a large set of diverse behaviors requires manual effort in the form of designing prompts.
On the other hand, Quality Diversity (QD) optimization has been shown to be capable of generating diverse Reinforcement Learning (RL) agent behavior.
In this work, we combine QD optimization with LLM-powered agents to iteratively search for prompts that generate diverse team behavior in a long-horizon, multi-step collaborative environment.
We first show, through a human-subjects experiment, that humans exhibit diverse coordination and communication behavior in this domain.
We then present a series of experiments showing that our approach captures behaviors that are difficult to observe without large-scale data collection, and a follow-up user study to show that these generated behaviors are human-like.
Our findings highlight the combination of QD and LLM-powered agents as an effective tool for studying teaming and communication strategies in multi-agent collaboration.

\end{abstract}

\section{Introduction}
\label{sec:introduction}

Robots and autonomous systems deployed in the real world must collaborate with and adapt to humans who exhibit diverse behaviors, expectations, and communication styles. 
For example, consider a robot that assists chefs in a restaurant kitchen. 
In terms of behavior, some chefs might want the robot to only chop vegetables, while others expect it to move ingredients to a chef working on a dish. 
In terms of communication, some chefs might give explicit orders to the robot, while others expect it to be more proactive and ask questions if necessary. 
For robots to adapt to such varied teams, they would need to build an understanding of how different humans might operate when performing these tasks.
As such, \emph{we address the problem of generating diverse, human-like teaming and communication behaviors in sequential decision-making tasks.} 

One approach to generating such teaming behaviors is through learning models of large-scale human data~\citep{carroll2019utility,pearce2023imitating}. 
However, collecting a sufficiently large and diverse dataset from collaborative domains, especially ones involving multiple interacting humans, is expensive and challenging~\citep{rogers2017research}. 

On the other hand, prior work has shown large language model (LLM)-powered agents to be a viable option for modeling human behavior~\citep{zhou2024sotopiainteractiveevaluationsocial,li2023theory,xie2024largelanguagemodelagents,yang2024oasis}.
With personality or strategy prompts to bias their actions, LLMs are shown to exhibit human-like behavior in social domains~\citep{park2023generative}.

\begin{figure*}
  \centering
  \includegraphics[width=1\textwidth]{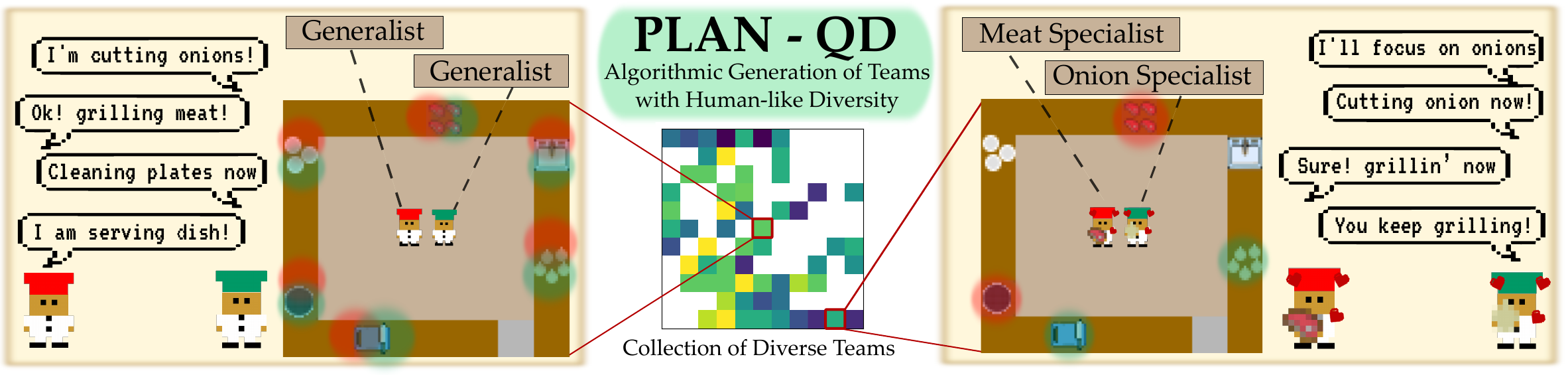}
  \caption{\planqd{} uses Quality Diversity (QD) optimization to generate a set of prompts to elicit human-like teaming diversity in LLM-powered agents. 
  The resulting teams exhibit distinct collaboration strategies (e.g., meat specialist with onion specialist), enabling the systematic study of communication and coordination in complex environments.} 
  \label{fig:hero}
\end{figure*} 

However, manually designing prompts to span the full range of possible behaviors is infeasible. To address the challenge of automatically generating prompts without collecting large-scale human teaming data, we turn to QD optimization~\citep{pugh2016quality}.
QD algorithms are designed to discover a set of high-performing solutions that are diverse with respect to specified measure functions. 
In the kitchen setting, an example measure function could count the number of ingredients picked, resulting in some agents never picking ingredients and focusing on other tasks, and others always picking them.

The key insight of our work is that \textit{QD optimization can be used to algorithmically generate prompts that elicit human-like teaming diversity in LLM-powered agents}. 
Starting from a basic prompt, our algorithm generates new prompts with the previous ones as stepping stones, iteratively creating a repertoire of behaviors along specified axes of diversity.
Thus, the manual effort for creating diverse behaviors is shifted from designing each individual prompt to simply specifying the required axes of diversity.

Through a human-subjects study, we show that \textit{human teams exhibit diversity along certain behavioral aspects} (e.g., workload distribution).
We generate diverse LLM-powered agents by using these behavioral aspects as axes of diversity, showing they replicate the effects of communication on human teamwork and exhibit human-like behavior.

We make the following contributions: 
(1) A human-subjects experiment that characterizes the diversity of teaming and communication behaviors in human teams in the domain Steakhouse~\citep{hsu2025fov}, inspired by the video game Overcooked~\citep{overcooked2,carroll2019utility}.
(2) \planqd{}, a novel framework for generating diverse LLM-agent prompts in collaborative multi-agent environments.
(3) A comparison showing that \planqd{} yields a broader and more diverse range of behaviors than standard LLM prompting (i.e., directly asking LLM to generate multiple prompts for diverse behavior).
(4) Empirical evidence that teams from \planqd{} are human-like in terms of behavior and the effect of communication on their teamwork.

\section{Background and Related Work}
\label{sec:background}

\emph{\textbf{Human Behavior Modeling.}}
Modeling human behavior~\citep{steinfeld2009oz,camerer2011behavioral} is an active research area in human-machine collaboration, drawing insights from human factors and cognitive science~\citep{salas2005there,hoffman2023inferring}. 
Particularly relevant to this work are studies on the effects of communication on team coordination and performance~\citep{mavridis2015review,tabrez2019explanation,natarajan2023human}.
However, these works lack effective models of human communication and variability~\citep{tabrez2020survey}.
Human modeling has also been viewed through the lens of zero-shot coordination, where agents are trained to collaborate with unseen partners.
Such approaches either augment human data via behavior cloning~\citep{carroll2019utility} or generative models~\citep{NEURIPS2021_ADAP,liang2024learning}, or generate purely synthetic agents~\citep{strouse2021collaborating}.
While these methods consider diversity in behaviors, they do not consider the role of communication.

Recent work has explored using LLMs to emulate human behaviors~\citep{zhou2024sotopiainteractiveevaluationsocial,li2023theory,xie2024largelanguagemodelagents,yang2024oasis}. 
However, these studies are mostly limited to text-based social settings, lacking embodied interaction with environments. 
On the other hand, LLM-powered agents in action-oriented domains rely on handcrafted prompts~\citep{zhang2024proagent,agashe2023llm}, restricting the extent to which their behavior can be systematically diversified.
In contrast, our work algorithmically generates personality prompts to elicit diverse behaviors in collaborative settings.

\emph{\textbf{Quality Diversity Algorithms and LLMs.}}
Quality Diversity (QD) algorithms search for a diverse collection of high-performing solutions~\citep{pugh2016quality}.
Typically, prior work has focused on QD optimization in search spaces with real numbers~\citep{lehman2011nslc,Cully2014RobotsTC,mouret2015illuminating,vassiliades2018line,Fontaine_2020,fontaine2021dqd,fontaine2022covariance}, including training diverse RL agents~\citep{cideron2020qdrl,nilsson2021pga,tjanaka2022approximating,batra2023proximal}.
However, their direct application to LLMs is challenging due to the large number of parameters in LLMs.

Hence, works applying QD optimization to LLMs search the space of prompts and leverage in-context few-shot prompting ~\citep{meyerson2024lmx,lim2024largelanguagemodelsincontext} or prompt mutation by another LLM~\citep{fernando2023promptbreeder,bradley2023quality,samvelyan2024rainbowteamingopenendedgeneration} to diversify LLM output.
The core idea of both types of approaches is the iterative improvement of prompts, with previously found prompts acting as stepping stones to find better prompts.
The efficacy of such iterative improvement has been shown previously in story generation and LLM red-teaming domains.
Our work leverages a similar insight and builds a framework that diversifies LLM-powered agents interfacing with low-level planners to generate diverse, collaborative behaviors in sequential decision-making tasks.

\section{Problem Definition}
\label{sec:problem}

We address the problem of finding diverse LLM-powered agents in collaborative sequential decision-making environments.
We formulate the environment as a decentralized Markov Decision Process (dec-MDP~\citep{bernstein2002complexity}) $\langle \gS, \gA, \gR, \gP, \gamma \rangle$ with $N$ agents, where $\gS$ is the state space, $\gA = \Pi_i^N \gA_i$ is the joint action space of all agents, $\gR: \gS \times \gA \rightarrow \sR$ is the common reward function that all agents receive, $\gP: \gS \times \gA \times \gS \rightarrow [0, 1]$ is the transition function, and $\gamma$ is the discount factor. 
The agents' goal is to maximize the discounted sum of rewards, $J=\Sigma_t \gamma^t r_t$, where $r_t$ is the reward obtained at timestep $t$.
With LLM-powered agents, the state and actions are provided and received via a text interface, with the space of textual inputs/outputs to the LLM defined by $\gT$.

\begin{figure*}
  \centering
  \includegraphics[width=0.9\textwidth]{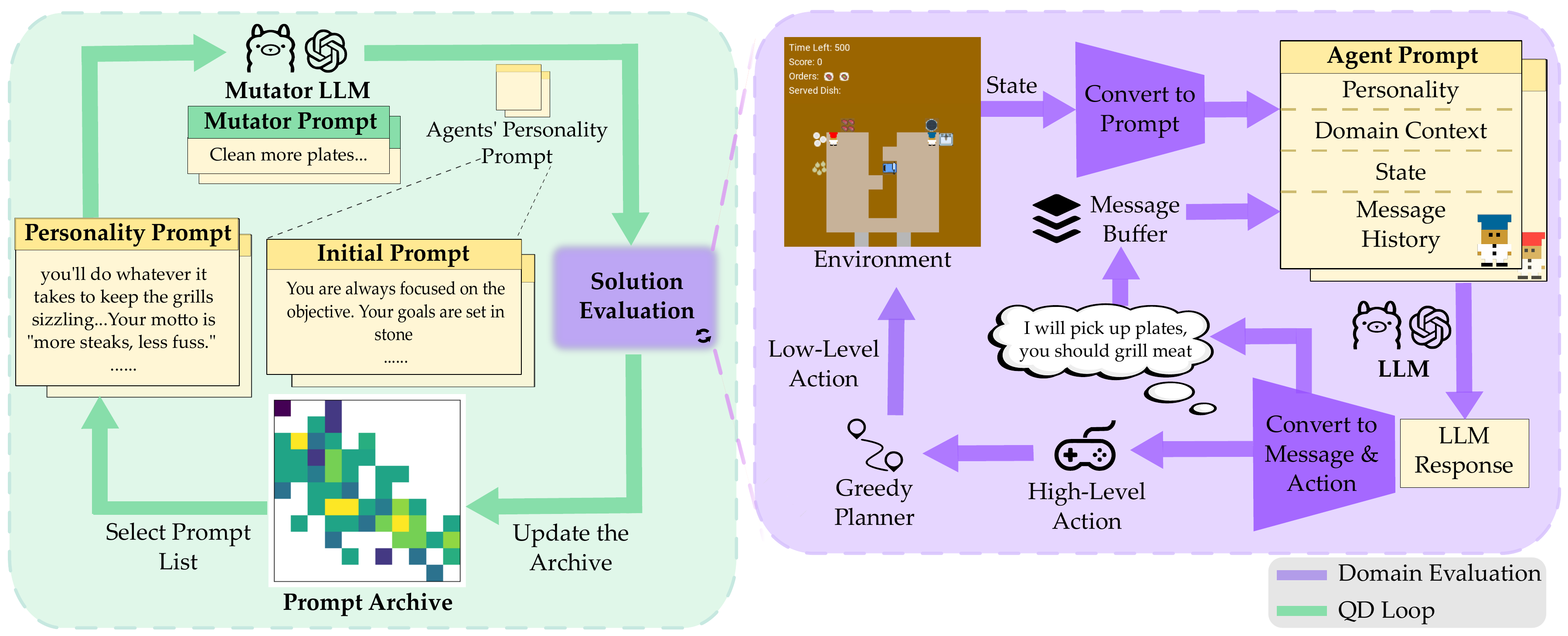}
  \caption{Overview of the \planqd{} framework, including the QD optimization (green arrows) and the LLM-powered agents.
  QD optimization repeatedly selects and mutates prompts to generate new prompts that are then evaluated in the environment (purple arrows).
  Only high-quality and diverse prompts are retained in the prompt archive.}
  \label{fig:qdllm}
\end{figure*}

To obtain diversity in agent behavior, we formulate the problem as \textbf{Quality Diversity optimization} applied to LLM-powered agents (QD-LLM).
Each LLM-powered agent receives additional instructions in the form of a prompt $x \in \gX \subseteq \gT$, resulting in a prompt list $\vx$ that defines the team. 
QD seeks to generate a diverse set of high-performing solutions (prompt lists in this context) by simultaneously optimizing for quality and behavioral diversity.
The diversity in individual or team behavior is characterized by a set of \textit{measure functions} $\vm: \gX^N \rightarrow \sR^k$, defining a \textit{measure space} $\gZ = \vm(\gX^N)$, while the \textit{quality} is captured by the objective function $J$, as defined earlier.
For example, in Steakhouse, one measure function could be the number of plates cleaned by the first agent.
The goal of QD-LLM is to discover prompt lists $\vx \in \gX^N$ such that $\vm(\vx) = \vz$ for all $\vz \in \gZ$, while maximizing $J$.
In the above example, solving the QD-LLM problem would result in a set of prompts that lead to high-performing teams where the first agent is diverse with respect to the number of plates cleaned.

Typically, the measure space is discretized into a finite number of cells, called the \textit{archive}, and QD-LLM's goal is to search for the best prompt list mapping to each cell. 
QD-LLM algorithms are evaluated using two metrics from the QD literature~\citep{pugh2016quality}: fraction of filled cells (\textit{coverage}), measuring the diversity of explored solutions, and the sum of objectives in all filled cells (\textit{QD-score}), quantifying the overall quality of solutions.

\section{Approach: The PLAN-QD Framework}
\label{sec:approach}

We propose \textbf{PLAN-QD} (\textbf{P}rompting \textbf{L}LM-powered \textbf{A}gents for \textbf{N}ovel Behavior via \textbf{Q}uality \textbf{D}iversity)  to solve the problem of generating diverse LLM-powered agents.
Our framework consists of the following components: (1) LLM-powered agents that include an interface between an LLM and the environment, along with a communication setup for agents to pass messages, and (2) QD optimization to find prompts that elicit diverse behavior in LLM-powered agents.
\fref{fig:qdllm} shows the overview of the complete framework.

\subsection{LLM-powered Agents}
\label{sec:llm_agent}

Our LLM-powered agents interface with an LLM (referred to as \emph{agent LLM}) to obtain both the actions to take in the environment and messages to communicate with other agents ({\color{Mulberry}\textbf{purple arrows}} in \fref{fig:qdllm}).

\noindent\textbf{LLM Input:}
The input to the LLM provides context, in the form of text, about the environment and its current state so that the LLM can make an informed choice about the next action to take.
Specifically, we prompt an LLM with the following information: 
(1) Personality to control the behavior of the agent;
(2) Context about the domain as a whole (e.g., goals, rules, etc.) to ground the LLM's decision-making;
(3) The text description of the current state 
to allow the LLM to react to changes in the environment;
(4) A limited history of the agent's previous actions and messages sent by all the agents, informing the LLM about what other agents are doing and what others might want it to do.
See \apref{app:agent_prompt} for an example of this query in the Steakhouse domain.

\noindent\textbf{LLM Output:}
The LLM outputs a high-level action and an optional message.
For example, in Steakhouse, a high-level action could be ``pick up plate'', and a corresponding message could be ``I will pick up plates, you should grill meat...''.
We pass the high-level action to a planner to convert it into a sequence of low-level actions (e.g., ``move left, up, and interact'').
We add the message to the buffer for future queries.

\noindent\textbf{Environment Simulation:}
At the beginning of an episode, we query the LLMs for all agents in a random sequence to obtain their high-level actions and messages.
The agents then step through the environment by taking the corresponding low-level actions provided by the planner.
Once the corresponding agent completes a high-level action or fails after a timeout, we re-query its LLM for a new high-level action and a message.

\subsection{QD Optimization}
\label{sec:qd_details}

To automate the process of finding personality prompts for diverse agents, PLAN-QD algorithmically searches for them using QD optimization ({\color{OliveGreen}\textbf{green arrows}} in \fref{fig:qdllm}) as follows:

\noindent\textbf{Prompt Selection:} 
\planqd{} maintains a prompt archive consisting of discretized cells, where each cell stores a list of high-quality personality prompts, one for each agent in the domain (two agents in Steakhouse), found during the optimization. 
At the start of optimization, the archive is empty, so an initial seed prompt is selected for all agents.
As the archive fills, \planqd{} selects a filled cell and samples its stored prompt list.
Since the archive contains high-quality prompt lists, the selected prompt list acts as a ``stepping stone'' for the algorithm to generate novel behavior.

\noindent\textbf{Prompt Mutation:} 
As defined in \sref{sec:problem}, the required axes of diversity are guided by a set of measure functions. 
The algorithm searches for prompts that promote diversity along these axes by querying a separate LLM (referred to as the \textit{mutator LLM} to differentiate from agent LLM) for new prompts.
The mutation process begins with the previously selected prompt list and leverages different forms of in-context learning~\cite{dong2022survey} to generate the new prompt list.
We propose the following three types of mutation, adapted from prior work~\cite{samvelyan2024rainbowteamingopenendedgeneration,meyerson2024lmx} to generate system prompts for agents:

\begin{enumerate}
   \item \textit{Directional}, the default mutation type, queries the mutator LLM to mutate the prompt list towards a random direction in the measure space (e.g., ``more number of plates cleaned''), generating a new prompt list that is expected to induce behaviors aligned with that direction.
    \item \textit{Language Model Crossover (LMX)} selects a second prompt list from the archive and queries the mutator LLM to generate a new prompt list that is a crossover of the two selected prompt lists (i.e., whose induced behavior is a mix of the two prompt lists).
    \item \textit{Random} is similar to ``Directional'', but does not specify a direction in the measure space, letting the mutator LLM use its internal notion of behavioral diversity to generate a new prompt list.
\end{enumerate}

See \apref{app:mutator_prompt} for example queries to the mutator LLM.

\noindent\textbf{Prompt Evaluation:} 
To evaluate the newly generated prompt list, \planqd{} simulates the corresponding LLM-powered agents in the environment as described in \sref{sec:llm_agent}.
The algorithm repeats the simulation multiple times and takes the median of the resulting objective and measure values to account for stochasticity.

\noindent\textbf{Archive Update:} 
The obtained measure values map the prompt list to a cell in the archive.
If that cell is empty or if the newly evaluated prompt list achieves a higher objective value than the one currently stored, the new prompt list replaces the existing one. 

By iteratively repeating \emph{prompt selection, mutation, evaluation, and archive update}, \planqd{} populates the archive with prompts that elicit high-quality and diverse behavior in LLM-powered agents.

\section{Experimental Validation}
\label{sec:experiment_validation}
Here, we describe the experimental setup, guided by the following motivations: (1) collect human data in a collaborative domain to establish a baseline for teaming behavior diversity, 
(2) evaluate whether our approach can replicate the diversity in human teaming behaviors, and
(3) investigate whether the generated behaviors are human-like.

\subsection{Domain}
\label{sec:domain}
\begin{figure}[bt!]
    \centering
    \subfloat[Open]{\includegraphics[width=0.11\textwidth]{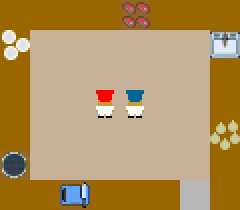} \label{fig:open_layout}}
    \hfill
    \subfloat[Ring]{\includegraphics[width=0.134\textwidth]{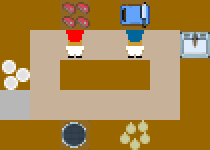} \label{fig:ring_layout}}
    \hfill
    \subfloat[Hallway]{\includegraphics[width=0.107\textwidth]{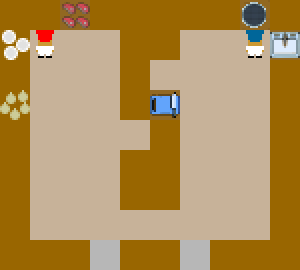} \label{fig:hallway_layout}}
    \hfill
    \subfloat[Forced Coordination]{\includegraphics[width=0.096\textwidth]{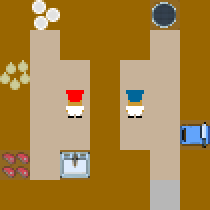} \label{fig:forced_layout}}
    \caption{Kitchen layouts in the Steakhouse environment. The four layouts span a spectrum from symmetrical (Open), where both players have equal access to all stations, to asymmetrical (Forced Coordination), where players depend on each other to access ingredients or complete tasks. Asymmetric layouts require more inter-agent coordination.}
    \label{fig:steakhouse_layouts}
\end{figure}

\noindent\textbf{Steakhouse Domain:} 
We chose a collaborative domain, \textit{Steakhouse}~\citep{hsu2025fov}, to test the efficacy of our algorithm in generating diverse LLM-powered agents.
Steakhouse, inspired by the game Overcooked~\citep{overcooked2} and its simulation environment~\citep{carroll2019utility}, introduces complex coordination challenges via larger and more varied layouts and multi-step recipes.
Efficient gameplay in this domain requires task division, coordination, and long-term planning. 
We used four distinct kitchen layouts: open, ring, hallway, and forced coordination (See \fref{fig:steakhouse_layouts}). These layouts were designed to represent a spectrum from symmetrical to asymmetrical ingredient accessibility, influencing how players divide tasks and rely on their partners. For example, the open layout allowed unrestricted movement and equal access to all kitchen components, whereas the forced coordination layout introduced strict dependencies, requiring players to pass ingredients across counters.

\noindent\textbf{User Study 1 (Teaming Data Collection):} 
To study how humans coordinate in this domain, we conducted a $2 \times 1$ between-subjects study ($n = 54$ participants; see \apref{app:user_study1} for the detailed study protocol) with two conditions: one where participants were allowed verbal communication via Zoom and another where they had no means of communication, requiring implicit coordination through gameplay.

\subsection{Measurements}
\label{sec:measures}

We defined individual and team behavior via the following measures (further details in \apref{app:measures}): %

\noindent\textbf{Teamwork Measures:}
We defined four quantitative measures of team performance and coordination: (1) \textit{Fitness}, i.e., discounted sum of rewards ($J$ in \sref{sec:problem}); (2) \textit{Average Action Delay}, i.e., number of timesteps between non-movement actions; (3) %
\textit{Percentage Contribution}, i.e., fraction of work done by the lowest contributor, with low values implying one player doing everything and high values implying equal division; 
(4) \textit{Specialization}, i.e., the extent to which each subtask is handled by a single player, with low values indicating evenly distributed tasks and high values indicating that players focused on specific subtasks.
Higher fitness and lower average action delay indicate better team performance, while varied percentage contribution and specialization indicate different coordination strategies.

We test the following hypothesis with the teamwork measures through the user study (\textbf{H1}):
\emph{Communication will affect teamwork measures in human teams.
Specifically, fitness will be higher with communication.}

\noindent\textbf{Workload Measures:}
We also defined a set of nine quantitative workload measures based on the differences in the number of times a player finishes a sub-task.
For example, ``Difference in Number of Onion Chopped'' counts the difference between the number of times the first player and the second player perform the sub-task of chopping onions. 
The nine sub-tasks we considered were: onions picked, onions placed on the board, onions chopped, meat picked, meat placed on the grill, dirty plates picked, clean plates picked, plates placed in the sink, and dishes served.
We used these measures as proxies for capturing diversity in teams of LLM-powered agents, as workload has been shown to be a strong indicator of team coordination behavior~\citep{gombolay2017computational,hoffman2019evaluating,fontaine2021importance}.
\subsection{QD Experiments}

We ran our experiments on four kitchen layouts (\sref{sec:domain}) with two conditions - with and without communication. 
We tested the following two algorithms for their effectiveness in generating a population of diverse agents, each given a budget of 100 prompt evaluations.

\noindent\textbf{\planqd{}:} 
Our framework, defined in \sref{sec:approach}, adapted to the Steakhouse domain.
We chose an initial pair of prompts and iteratively mutated them until the budget was exhausted.
We used three mutation types (see \sref{sec:qd_details}): \planqd{} (Directional), \planqd{} (LMX), and \planqd{} (Random), with \planqd{} (Directional) serving as the default.

We maximize ``Fitness'' (objective function for QD) and diversify ``Difference in Number of Meat Put on Grill'', ``Difference in Number of Dish Served'' and ``Difference in Number of Onion Chopped'' (measure functions for QD).

\noindent\textbf{Random Batch:}
A baseline in which an LLM generated 100 prompts to evaluate based on the same initial pair of prompts as in \planqd{}.
The key differences here are that the mutator LLM does not have any explicit measures to diversify and does not have stepping stones of prompts that led to diverse and high-quality agents.

To account for stochasticity, each evaluation reports the median objective and measures over four repetitions of a 500-timestep gameplay episode, where the agent interacts with the environment.
\apref{app:algos} contains other hyperparameter values of \planqd{} and Random Batch.

We test the following hypothesis with QD experiments (\textbf{H2}):
\emph{\planqd{} will have greater archive coverage (i.e., capture more diverse behaviors) compared to Random Batch, due to its iterative improvement of prompts.}

\subsection{Human-likeness Experiments}

We test the human-likeness of \planqd{}'s agents through two experiments. 

\noindent\textbf{Trend Comparison:}
First, to study if the effect of communication is human-like, we compared the trends in teamwork measures between with and without communication conditions in human teams, followed by the same comparison in teams generated by \planqd{}.
We test the following hypothesis (\textbf{H3}):
\emph{Behavioral trends between communication conditions exhibited by \planqd{}'s agents will match those observed in human teams, as we explicitly diversify over the same measures along which human teams show variation.}

\noindent\textbf{User Study 2 (Human-likeness Evaluation):} 
Second, to study whether the behavior (without communication) is human-like, we conducted a second user study ($n = 40$ participants; see \apref{app:user_study2} for the detailed study design).
We showed participants three types of videos: human gameplay, \planqd{}'s agents, and heuristic agents (i.e., AI agents that choose high-level actions based on some heuristics). 
Participants were asked to (1) classify each video as coming from a ``human'' or ``AI''  team, and (2) rate the human-likeness of the behavior on a 7-point Likert scale.
We test the following hypothesis (\textbf{H4}):
\emph{The behavior of \planqd{}'s agents will be classified as more human-like compared to the behavior of heuristic agents, due to the inherent human-like decision making of LLM-powered agents~\cite{park2023generative}.}

\section{Result and Discussion}
\label{sec:results}

\begin{table*}[ht!]
    \centering
    \footnotesize
    \begin{tabular}{lcccccccc}
        \toprule
        & \multicolumn{2}{c}{Open} & \multicolumn{2}{c}{Ring}
        & \multicolumn{2}{c}{Hallway} & \multicolumn{2}{c}{Forced Coordination} \\
        \cmidrule(lr){2-3} \cmidrule(lr){4-5} \cmidrule(lr){6-7} \cmidrule(lr){8-9}
        & w/ Comm. & w/o Comm. & w/ Comm. & w/o Comm. & w/ Comm. & w/o Comm. & w/ Comm. & w/o Comm. \\
        \midrule

        \textbf{Random Batch} & $0.22_{0.07}$ & $0.25_{0.07}$ & $0.25_{0.04}$ & $0.24_{0.04}$ & $0.22_{0.04}$ & $0.20_{0.03}$ & $0.04_{0.02}$ & $0.05_{0.02}$ \\

        \textbf{PLAN-QD} & & & & & & & & \\
               
        \textbf{\quad(Random)} & $0.22_{0.07}$ & $0.21_{0.07}$ & $0.23_{0.03}$ & $0.22_{0.04}$ & $0.20_{0.03}$ & $0.19_{0.04}$ & $0.04_{0.02}$ & $0.04_{0.02}$ \\
        
        \textbf{\quad(LMX)} & $0.24_{0.07}$ & $0.21_{0.06}$ & $0.20_{0.02}$ & $0.22_{0.03}$ & $0.21_{0.03}$ & $0.19_{0.04}$ & $0.03_{0.02}$ & $0.03_{0.02}$ \\

        \textbf{\quad(Directional)} & $\textbf{0.32}_{0.12}$ & $\textbf{0.33}_{0.11}$ & $\textbf{0.31}_{0.06}$ & $\textbf{0.32}_{0.06}$ & $\textbf{0.30}_{0.07}$ & $\textbf{0.33}_{0.07}$ & $\textbf{0.07}_{0.03}$ & $\textbf{0.11}_{0.03}$ \\

        \bottomrule
    \end{tabular}
    \caption{Averaged archive coverage values across $12C2 = 66$ distinct pairs of measures. Standard deviations across the pairs are reported in subscripts. PLAN-QD with directional mutation is consistently better than other baselines.}
    \label{tab:qd_versus_random_combined_flipped}
\end{table*}

\begin{table}
    \centering
    \small
    \renewcommand{\arraystretch}{1.2}
    \setlength{\tabcolsep}{0pt}

    \begin{tabular}{lcccc}
        \toprule
        \textbf{Metric} & \textbf{Human} & \textbf{\planqd{}} & \textbf{Human} & \textbf{\planqd{}} \\
        \midrule
        & \multicolumn{2}{c}{\textbf{Open}} & \multicolumn{2}{c}{\textbf{Ring}} \\
        
        \midrule

        Fitness 
        & \colorbox{lime!30}{$\boldsymbol{-2.71\%}$} & \colorbox{lime!30}{$\boldsymbol{-5.00\%}$}
        & \colorbox{lime!30}{$\boldsymbol{-4.33\%}$} & \colorbox{lime!30}{$\boldsymbol{-6.88\%}$} \\

        Avg. Action Delay
        & \colorbox{lime!30}{$\boldsymbol{+4.65\%}$} & \colorbox{lime!30}{$\boldsymbol{+1.20\%}$}
        & \colorbox{red!20}{$\boldsymbol{+7.39\%}$} & \colorbox{red!20}{$\boldsymbol{-3.63\%}$} \\

        Percent Contrib.
        & \colorbox{lime!30}{$\boldsymbol{-3.78\%}$} & \colorbox{lime!30}{$\boldsymbol{-2.25\%}$}
        & \colorbox{lime!30}{$\boldsymbol{+7.73\%}$} & \colorbox{lime!30}{$\boldsymbol{+9.33\%}$} \\

        Specialization
        & \colorbox{lime!30}{$\boldsymbol{+4.21\%}$} & \colorbox{lime!30}{$\boldsymbol{+2.40\%}$}
        & \colorbox{red!20}{$\boldsymbol{-7.31\%}$} & \colorbox{red!20}{$\boldsymbol{+4.21\%}$} \\

    \midrule

        & \multicolumn{2}{c}{\textbf{Hallway}} & \multicolumn{2}{c}{\textbf{Forced Coordination} } \\
        \midrule

        Fitness
        & \colorbox{lime!30}{$\boldsymbol{+0.99\%}$} & \colorbox{lime!30}{$\boldsymbol{+1.57\%}$}
        & \colorbox{lime!30}{$\boldsymbol{+24.6\%}$} & \colorbox{lime!30}{$\boldsymbol{+37.2\%}$} \\

        Avg. Action Delay
        & \colorbox{red!20}{$\boldsymbol{-6.55\%}$} & \colorbox{red!20}{$\boldsymbol{+4.13\%}$}
        & \colorbox{lime!30}{$\boldsymbol{-20.7\%}$} & \colorbox{lime!30}{$\boldsymbol{-2.01\%}$} \\

        Percent Contrib.
        & \colorbox{lime!30}{$\boldsymbol{+8.43\%}$} & \colorbox{lime!30}{$\boldsymbol{+0.96\%}$}
        & \colorbox{lime!30}{$\boldsymbol{+19.8\%}$} & \colorbox{lime!30}{$\boldsymbol{+8.23\%}$} \\

        Specialization
        & \colorbox{red!20}{$\boldsymbol{-1.77\%}$} & \colorbox{red!20}{$\boldsymbol{+0.72\%}$}
        & \colorbox{lime!30}{$\boldsymbol{+9.38\%}$} & \colorbox{lime!30}{$\boldsymbol{+8.91\%}$} \\
        \bottomrule
    \end{tabular}

    \caption{Percentage difference in teamwork measures (\sref{sec:measures}) with communication for human and \planqd{} teams, with positive difference indicating a higher metric value with communication. 
    Trends for \planqd{} match those of humans in 12/16 cases ({\color{lime!40!black}\textbf{colored green}}).
    }
    \label{tab:comparison}
\end{table}

\subsection{Humans Exhibit Diverse Behaviors Depending on Communication and Layout}
\label{ssec:results_user_study}

To test \textbf{H1} on collected teaming data (User Study 1), we conducted independent t-tests comparing fitness between the ``with communication'' and ``without communication'' conditions within each layout. 
The t-tests showed no significant differences because of the variance among participants.
However, trends suggest that communication improved performance in more asymmetric layouts (e.g., forced coordination) but had little or negative effects in more symmetric layouts (e.g., open, ring). 
Notably, forced coordination, which required strong role differentiation, showed the greatest performance improvement with communication, likely because verbal coordination helped players manage dependencies more effectively. 
Qualitatively, we also noted that teammates frequently assisted struggling participants by actively communicating and providing help in this layout.

Communication also influenced collaboration styles. 
For example, percent contribution (i.e., cooperating effort) increased with communication in all layouts except open, where task division was already balanced. 
Interestingly, specialization (i.e., role division) increased in the open layout, suggesting that communication helped with strategic task division (see \tref{tab:comparison} for trends).
\vbnote{We should probably move the table near this instead of keeping it in Sec. 6.3}

Additionally, we found a significant difference in fitness when considering both layout and communication as factors, indicating that spatial constraints influenced coordination strategies. 
Similarly, other teamwork measures (e.g., specialization) also showed significant effects. 
We further observed that individual differences, such as background knowledge and personality, influenced teaming behavior with communication.
We provide additional details about the observations from the user study in \apref{app:user_study_res}.

Qualitative responses further supported these findings. 
Participants in the ``without communication'' condition reported coordination difficulties and expressed a preference for having some form of communication, in their exit interviews. 
In summary, while communication generally improved performance in forced coordination, the impact was weaker in layouts where tasks could be executed more independently. 
\emph{Our results suggest that communication effects should be studied alongside both spatial task constraints (which shape task roles) and individual skill and personality differences}, aligning with findings in human factors research~\citep{driskell2006makes,hays2022tale}.

\subsection{PLAN-QD's Agents are Diverse}
\label{ssec:qdllm_benchmark}

To test \textbf{H2}, we adapted the coverage metric (\sref{sec:problem}), defined as the number of cells filled by the algorithm in the archive (the discretized measure space). 
However, we have 9 workload and 3 teamwork measures (excluding fitness), and each algorithm only evaluates 100 prompts.
Hence, the 12-dimensional measure space will be mostly empty, and comparing the coverage there will not be very informative.
Thus, we looked at two-dimensional planes defined by distinct pairs of measures, resulting in $C(12,2)=66$ coverage values, and report the average in \tref{tab:qd_versus_random_combined_flipped}.

\planqd{} with directional mutation significantly outperformed other baselines, based on a one-sample test of proportions (higher coverage in all eight layout–condition pairs; a greater proportion than the expected random proportion of $0.50, p = .004)$, \textbf{validating H2}.
We present QD-score comparisons in \apref{app:coverage_res}.
\planqd{} with other mutation types obtain lower coverage, \emph{highlighting the importance of specifying the behavioral direction for effective prompt mutation.}

\begin{figure*}
    \centering
    \begin{subfigure}[t]{0.495\textwidth}
        \includegraphics[width=1\linewidth]{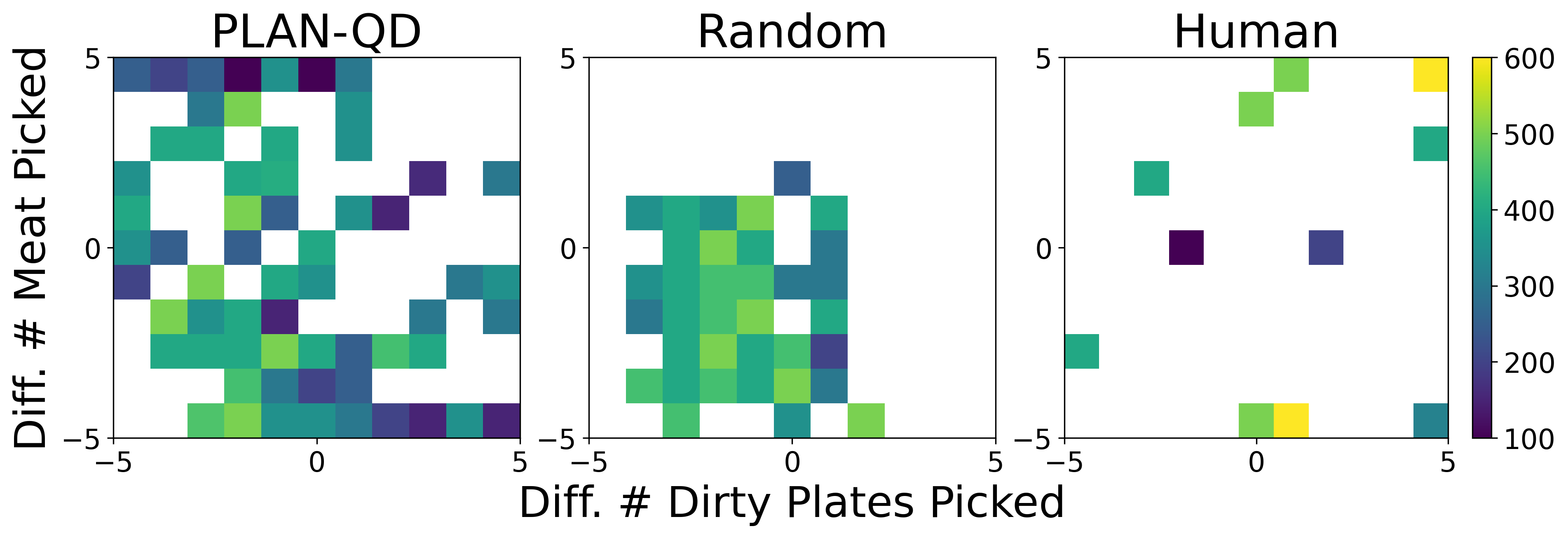} 
        \captionsetup{justification=centering}
        \caption{Workload Measures on Open Layout \textbf{with communication}.}
    \label{fig:qdllm_random_human_comp1}
    \end{subfigure}
    \begin{subfigure}[t]{0.495\textwidth}
        \includegraphics[width=1\linewidth]{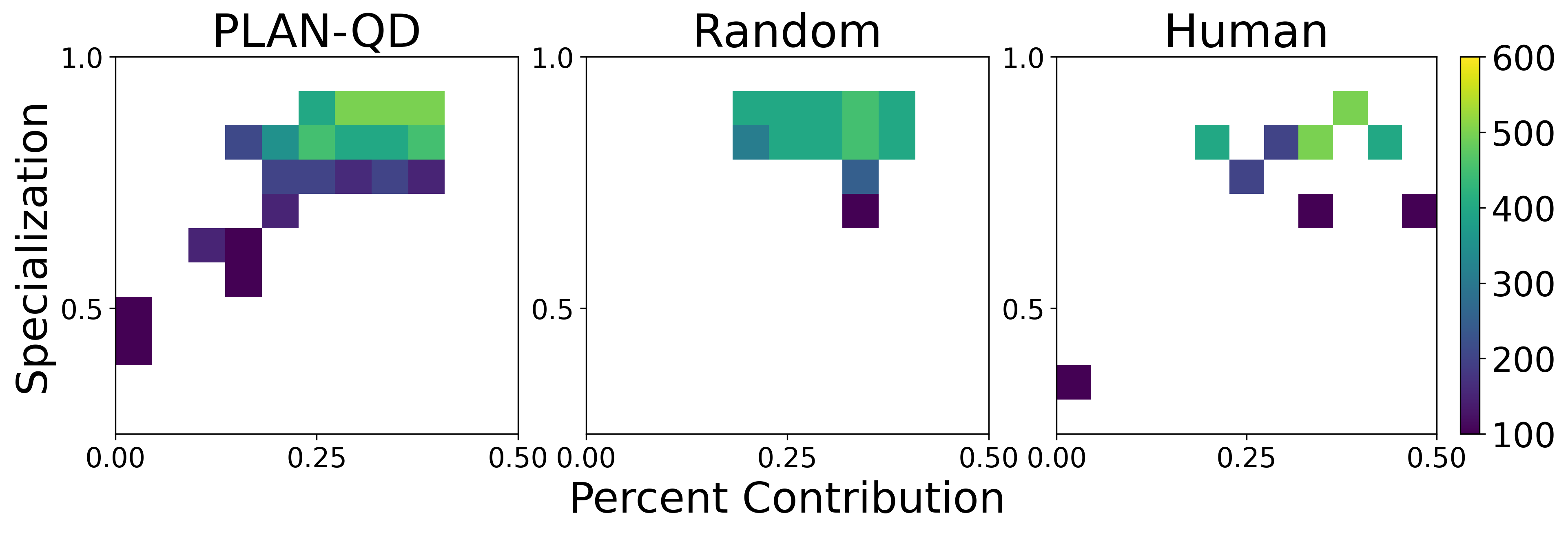}
        \captionsetup{justification=centering}
        \caption{Teamwork Measures on Forced Coordination Layout \textbf{without communication}.}
    \label{fig:qdllm_random_human_comp2}
    \end{subfigure}

    \caption{Example archive heatmaps from human data, \planqd{}, and Random Batch, colored by fitness value. 
    \planqd{} generates agents covering a wider range of behavior compared to Random Batch, including certain extremes observed in the human data.
    \textbf{Videos} of example behaviors are included in the supplementary material.}
    \label{fig:qdllm_random_human_comp}
\end{figure*}

To analyze the coverages qualitatively, we plot heatmaps of the prompt archive in \fref{fig:qdllm_random_human_comp}.
\planqd{} explicitly diversified both measures in \fref{fig:qdllm_random_human_comp1}, and we clearly see more cells being filled in the archive compared to Random Batch.
In fact, some of the extreme behavior seen in human data (e.g., high positive difference in meat picked) is only exhibited by agents found by our method.
\planqd{} also finds certain behaviors (e.g., high negative difference in meat picked) that are not found in human data, \emph{highlighting its benefit in augmenting human datasets.}
Furthermore, even when looking at measures that were not explicitly diversified by PLAN-QD (\fref{fig:qdllm_random_human_comp2}), we see a better coverage than Random Batch, including certain rare behaviors such as low percent contribution and specialization similar to those exhibited by human users who did not fully understand the game rules.%

\noindent\textbf{PLAN-QD Mutation Generalizes Across Language Models:}
To study the effect of specific language models in the \planqd{} framework, we ran two ablations: various mutator LLMs with agent LLM fixed, and various agent LLMs with a fixed personality prompt.
\tref{tab:planqd_directional_different_models} shows the averaged coverage values across measure pairs for the open and forced coordination layouts using different mutator models.
We observe similar coverage across models, which we attribute to the iterative prompt improvement process underlying \planqd{}.
However, in the agent LLM ablation, the models' capability heavily impacted the peak performance of the agents.
See \apref{app:agent_llm_ablations} for additional analysis.

\begin{table}
    \centering
    \footnotesize
    \begin{tabular}{lcc}
        \toprule
        \textbf{Model} & \textbf{w/ Comm.} & \textbf{w/o Comm.} \\
        \midrule
        \multicolumn{3}{c}{\textbf{Open}} \\
        \midrule
        LLaMA 3.1 70B Instruct & $0.32_{0.12}$ & $0.33_{0.11}$  \\
        Qwen 2.5 72B Instruct & $0.34_{0.10}$ & $0.35_{0.09}$  \\
        Deepseek Chat V3 & $\textbf{0.35}_{0.10}$ & $\textbf{0.39}_{0.08}$ \\
        OpenAI GPT-4.1 nano & $0.34_{0.10}$ & $0.34_{0.09}$ \\
        Gemini 2.5 Flash & $0.33_{0.09}$ & $0.34_{0.09}$ \\
        \midrule
        \multicolumn{3}{c}{\textbf{Forced Coordination}} \\
        \midrule
        LLaMA 3.1 70B Instruct & $0.07_{0.03}$ & $0.11_{0.03}$ \\
        Qwen 2.5 72B Instruct & $0.11_{0.04}$ & $0.14_{0.04}$ \\
        Deepseek Chat V3 & $0.13_{0.05}$ & $\textbf{0.15}_{0.05}$ \\
        OpenAI GPT-4.1 nano & $\textbf{0.14}_{0.05}$ & $\textbf{0.15}_{0.05}$ \\
        Gemini 2.5 Flash & $\textbf{0.14}_{0.05}$ & $\textbf{0.15}_{0.05}$ \\
        \bottomrule
    \end{tabular}
    
    \caption{Averaged archive coverage with \planqd{} (Directional) and various mutator LLMs. 
    Standard deviations are reported in subscripts. 
    All mutator LLMs result in similar coverage due to \planqd{}’s iterative improvement process.
    }
    \label{tab:planqd_directional_different_models}
\end{table}

\subsection{PLAN-QD's Agents are Human-like} 
\label{ssec:qdllm_trends}

To test \textbf{H3}, we used human team data from User Study 1 and compared it with teams generated by \planqd{}.
\tref{tab:comparison} shows the trends in teamwork measures between with and without communication conditions for the corresponding teams.
We find that \planqd{}'s agents follow trends observed in human users, via a one-sample test of proportions (12 out of 16 layout–metric combinations matched human trends; a greater proportion than the expected random proportion of $0.50, p = .038$). 
Notably, our approach successfully replicates trends in fitness and percent contribution trends. 
However, our approach fails to match action delay and specialization trends in the ring and the hallway layouts. 
We hypothesize this discrepancy is caused by factors such as differences in background knowledge of the users. 
Thus, these results \textbf{partially validate H3}.

To test \textbf{H4}, we used responses from User Study 2, where participants were asked to classify and rate human-likeness from videos of human teams, \planqd{} agents, and heuristic agents.
We ran a pairwise $\chi^2$ proportion test with Holm--\v{S}id\'{a}k p-value correction on the binary classification responses, and a one-way ANOVA followed by Tukey's HSD on the human-likeness Likert scores. 
We first observed that both binary classification and human-likeness scores were significantly different for human and heuristic videos ($p = .009$ and $p < .001$, respectively), validating that users could sufficiently distinguish human-like from non-human-like behavior in our study.
We observed no significant difference between \planqd{} and heuristic videos ($p=.118$) in terms of binary classification.
However, one-way ANOVA showed a significant effect of video type on the human-likeness score ($F(2,393)=9.17;p<.001$), with post-hoc Tukey's HSD test showing \emph{\planqd{} videos to be significantly more human-like than heuristic videos} (mean difference of .80 with $p=.003$), \textbf{partially validating H4}.

\section{Discussions and Conclusions}
\label{sec:conclusion}

\noindent\textbf{Limitations:}
One of the limitations of \planqd{}'s agents is their high computational cost due to evaluations requiring multiple LLM queries. 
To address this limitation, a promising approach is distilling specialized models, which would retain key behavioral properties while being significantly less compute-intensive~\citep{hinton2015distillingknowledgeneuralnetwork, zhao2023survey}. Alternatively, one could explore generating policy code for agents instead of querying LLMs directly for actions~\citep{wang2024executable}. However, incorporating communication strategies and personality within such a framework remains a challenging open problem.

Furthermore, our LLM-powered agents' communication is intertwined with action selection, which limits the natural flow of interaction.  
Future work should explore communication frameworks that incorporate models from human teaming research, determining when and how to communicate~\citep{vasil2020world,van2021team}. 

Finally, in our framework, LLM-powered agents only select high-level actions. 
Allowing LLMs to directly choose low-level actions will enable more diverse collaboration strategies, but would significantly increase compute time due to more frequent LLM queries.
Moreover, current LLMs struggle with planning and reasoning in low-level action spaces~\citep{valmeekam2023planning,tamkin2021understanding}.

\noindent\textbf{Conclusions:}
In this work, we address the challenge of algorithmically generating teaming and communication behaviors with human-like diversity using foundation models. 
We propose \planqd{}, a novel framework that applies Quality Diversity (QD) optimization to algorithmically generate diverse LLM agents capable of communication. 
We demonstrate that:
(1) humans exhibit diverse teaming behaviors, affected by communication, through a human-subjects study,
(2) \planqd{} covers a broader range of diverse and extreme human behaviors that are often difficult to observe in limited human trials, and
(3) teams generated by \planqd{} exhibit human-like behavior as well as being affected by communication in the same way as human teams. 
Our findings highlight the advantages of leveraging LLMs and QD to generate agents that can represent multi-human teams, providing a framework for studying human-AI teams.

\section*{Acknowledgments}
This work was supported by NSF CAREER \#2145077 and DARPA EMHAT HR00112490409. 
We would like to thank Ryan Boldi and Dhruv Kaul for their early work on prompt design for LLM-powered agents. Additionally, we would also like to thank Dana Hughes, Lucy Romero, and Simon Stepputtis for their contributions to the development of the preliminary user study design.

\bibliography{reference}

\clearpage
\appendix

\section{Steakhouse Domain}
\label{app:domain}

\begin{figure}
    \centering
    \begin{subfigure}[t]{0.95\columnwidth}
        \includegraphics[width=1\linewidth]{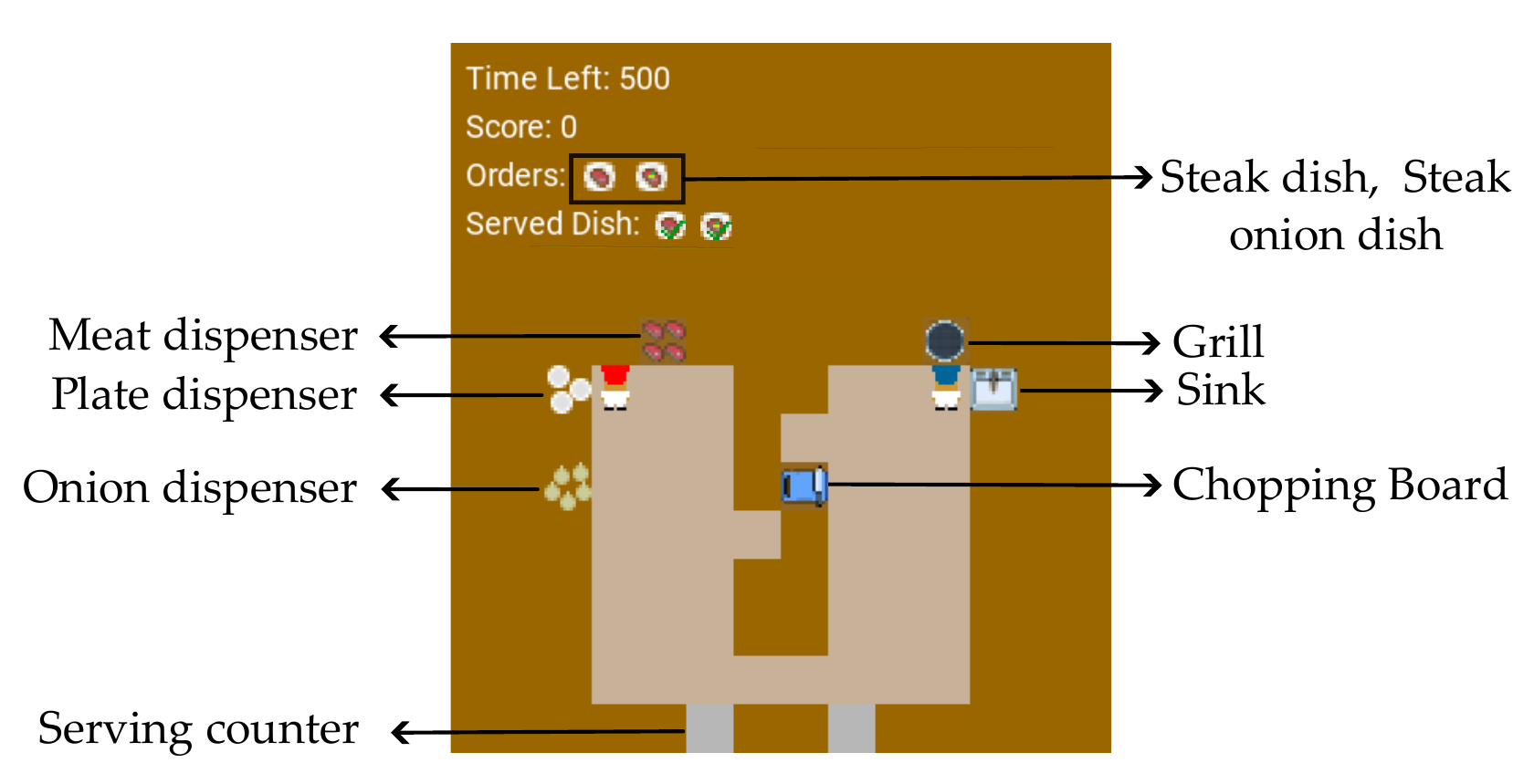} 
        \captionsetup{justification=centering}
    \label{fig:steakhouse_details}
    \end{subfigure}
    \begin{subfigure}[t]{0.95\columnwidth}
        \includegraphics[width=1\linewidth]{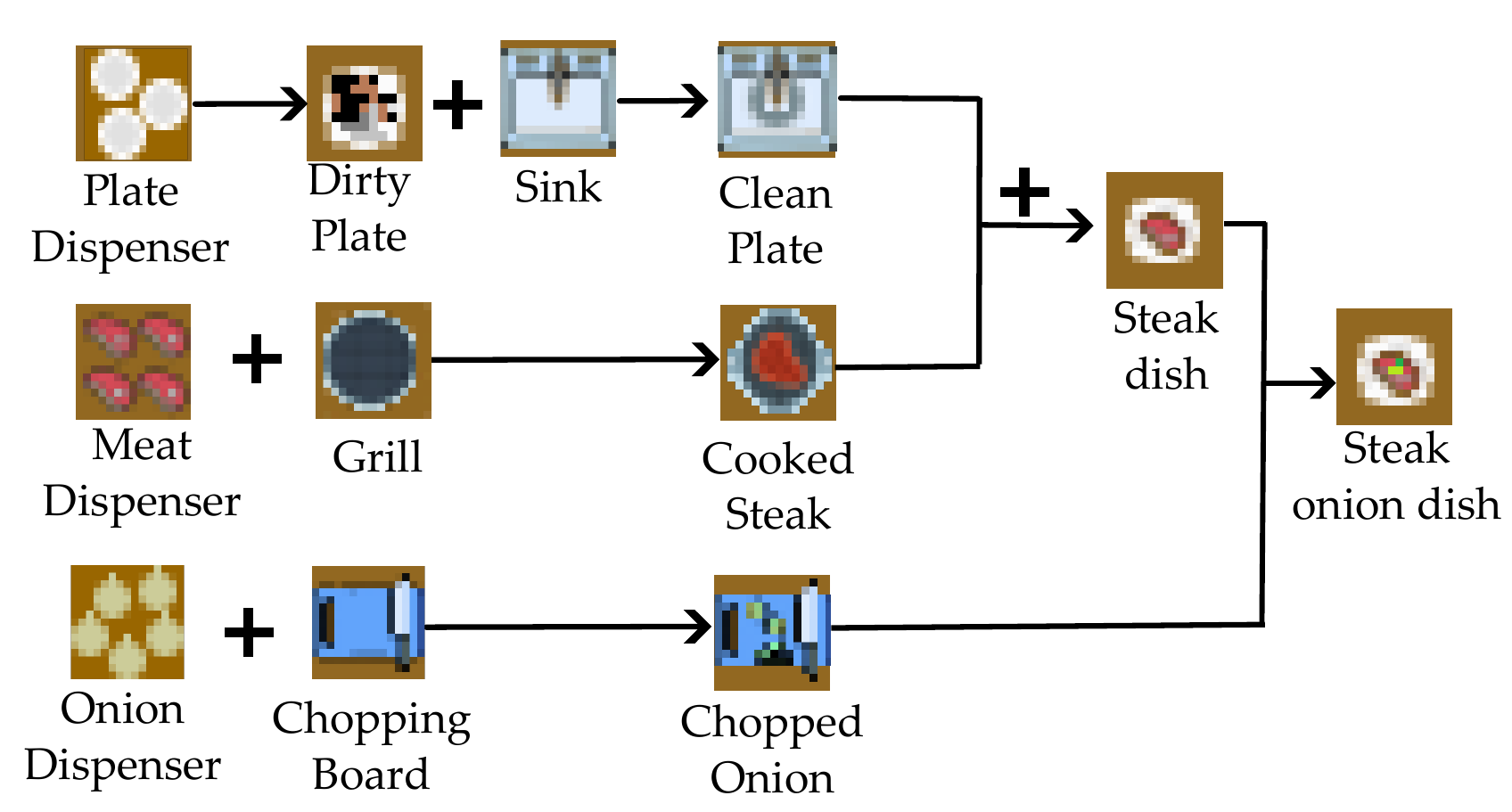}
        \captionsetup{justification=centering}
    \label{fig:process}
    \end{subfigure}

    \caption{Overview of the Steakhouse domain with an example environment on the left and the recipes on the right.}

    \label{fig:domain_details}
\end{figure}

Steakhouse is a collaborative cooking domain involving multiple agents (two in our experiments).
The agents need to prepare the requested dishes and deliver as many of them as possible within a time limit of 500 timesteps.
The agents operate in a fully observable grid (\fref{fig:domain_details}), and are controlled via six actions: North, South, East, West (move up, down, left, or right, respectively), Stay (remain in the same position), and Interact (engage with the environment). At each timestep, agents can only perform one of these six actions.

The environment contains three types of items: raw meat, dirty plate, and raw onion, each with a corresponding dispenser that contains an infinite amount of the particular resource.
Agents can pick up the ingredients by interacting with their dispensers.

There are also three appliances in the kitchen: a grill, a sink, and a chopping board. 
Meat can be placed on the grill to turn into cooked meat after 60 timesteps,
a dirty plate can be cleaned by interacting with the sink thrice, and a raw onion can be chopped by interacting with the chopping board twice.

The possible dish requests include two recipes: ``steak'' and ``steak with onion''. 
``Steak'' requires cooked meat and a clean plate, whereas ``steak with onion'' additionally requires a chopped onion.
At any given time, players will have two orders in the order list. 
The first two orders are always ``steak'' and ``steak with onion''.
Subsequent orders are selected from the two options uniformly randomly when one of the orders is delivered.
Delivering dishes in the correct order of the order list obtains 100 points per dish, whereas delivering out of the order of the order list only obtains 20 points.

\section{Algorithm Details}

\subsection{Pseudocode}
\label{app:pseudocode}

\begin{algorithm}
\LinesNumbered
\DontPrintSemicolon
\caption{QD optimization to generate prompts for diverse agents}
\label{alg:planqd}
\KwIn{
    Initial prompt $x^{(0)}$, max iterations $N_{iter}$, batch size $N_B$, number of repeats $N_{repeat}$
}
\KwOut{
    Archive of diverse prompt lists $\gA_{QD}$
}

Initialize $\gA_{QD}$ \;

\For{$i \in \{1, 2, \dots, N_{iter}\}$}{
    \For{$j \in \{1, 2, \dots, N_B\}$}{
        \If(\tcp*[f]{Prompt selection}){$\gA_{QD} = \phi$ \label{alg:sel_start}}{
            $\vx \leftarrow [x^{(0)}, x^{(0)}]$ \label{alg:init}
        }
        \Else{
            $\vx \leftarrow$ Uniformly select from $\gA_{QD}$ \label{alg:post_init}
        } \label{alg:sel_end}
        $\vx' \leftarrow mutate(\vx)$ \tcp*{Prompt mutation} \label{alg:mut}
        $J_{1 \dots k}, \vm_{1 \dots k} \leftarrow$ Evaluate $\vx$ in Steakhouse $N_{repeat}$ times \tcp*{Prompt evaluation} \label{alg:eval_start}
        $J \leftarrow median(J_{1 \dots k})$ \;
        $\vm \leftarrow median(\vm_{1 \dots k})$ \label{alg:eval_end} \;
        Update $\gA_{QD}$ \tcp*{Archive update} \label{alg:arch}
    }
}
\end{algorithm}

\aref{alg:planqd} provides the pseudocode of the QD optimization part of \planqd{} framework.
The QD loop is inspired by MAP-Elites~\citep{Cully2014RobotsTC,mouret2015illuminating} and consists of four steps: prompt selection, prompt mutation, prompt evaluation, and archive update.

For the prompt selection step (Lines~\ref{alg:sel_start}-\ref{alg:sel_end}), the algorithm selects a prompt list uniformly from the archive (Line~\ref{alg:post_init}).
However, in the first iteration, when the archive is empty, copies of the initial prompt $x^{(0)}$ are selected instead (Line~\ref{alg:init}).

Then, for each personality prompt in the selected prompt list, the algorithm creates a random mutation direction and provides this direction as a language direction via a mutation prompt.
The mutator LLM mutates each personality prompt independently based on the provided direction, resulting in a new prompt list (Line~\ref{alg:mut}).

The algorithm evaluates the new prompt list $N_{repeat}$ times and takes the median objective and measure values (Lines~\ref{alg:eval_start}-\ref{alg:eval_end}).
If the cell in the archive that the new prompt list is mapped to is empty or contains a prompt list with a lower objective, then the new prompt list is added to the archive (Line~\ref{alg:arch}).                 
                         
\subsection{Prompts for LLM-powered Agent}
\label{app:agent_prompt}

The prompts for LLM-powered agents contain context about the domain, the current state of the environment, and, optionally, messages communicated between the agents.
We provide an example of the agent prompt structure used in our experiments in the Steakhouse domain (\fref{fig:agent_state_prompt}).

First, we provided the agent with their name and the name of the other agent. The first agent was always named Alice, and the second Bob. 
We then provided the agents with a brief textual description of the domain, which explained what appliances were in the environment and any layout-specific constraints. 
For example, in the forced coordination layout, we described how the agents were separated from each other.
We then provided the agents with a textual description of the game, including the rules and recipe requirements.

In the textual description, we explained the purpose of counters and distinguished between two types of counters that we defined: \textit{shared counters} and \textit{general counters}.
Shared counters were designated counters that we manually chose as important for collaboration. For example, in the forced coordination environment, only three of the counters can be used for sharing between the two sections. Such counters also exist in the hallway and ring layouts, but not in the open layout. 
Counters not specially designated like this were labeled as general counters.
Although this distinction was made to give the agents more functionality to collaborate with each other, the agents were not explicitly enforced or encouraged to use either type of counter during gameplay.

In the state, we provided the following content in the prompt:
\begin{itemize}
    \item \textbf{Inventory: } what both agents are holding (e.g., Alice is holding a dirty plate)
    \item \textbf{Environment details: } the status of the appliances in the layout (e.g., what is on the grill) 
    \item \textbf{Location info: } where the appliances are located in the layout (e.g., the grill is 10 units away)
    \item \textbf{Order list: } list of current orders for the agent to deliver.
    \item \textbf{Action history: } limited history of past actions taken by the agent.
    \item \textbf{Message history: } limited history of past messages sent by the agent. 
\end{itemize}

Finally, we provided the agent with their objective and an explanation of the rewards obtained from delivering dishes (outlined in \apref{app:domain}). We then asked the agent to select an action from a list of available high-level actions. We provided the agent with a filtered list of actions based on what was possible/available in the current state. For example, if the agent was holding a meat, we did not provide them with any high-level actions regarding picking up an ingredient. Below is a list of all high-level actions the agent could choose from, organized by category:

\noindent \textbf{Counter Actions:}
\begin{itemize}
    \item ``Pick up [item] from [counter]''
    \item ``Place [item] on nearest general counter''
    \item ``Place [item] on [shared counter]''
\end{itemize}

\noindent \textbf{Plate Actions:}
\begin{itemize}
    \item ``Pick up a dirty plate from the dirty plate dispenser''
    \item ``Place dirty plate in hand in [sink]''
    \item ``Do one rinse of the dirty plate in [sink]''
    \item ``Pick up the clean plate from [sink]''
    \item ``Use clean plate in hand to pick up steak from [grill]''
\end{itemize}

\noindent \textbf{Onion Actions:}
\begin{itemize}
    \item ``Pick up an onion from the onion dispenser''
    \item ``Put raw onion in hand on [chopping board]''
    \item ``Do one chop of the onion on [chopping board]''
    \item ``Add garnish from [chopping board] to the steak dish in hand''
\end{itemize}

\noindent \textbf{Meat Actions:}
\begin{itemize}
    \item ``Pick up a meat from the meat dispenser''
    \item ``Put raw meat in hand on [grill] to cook''
    \item ``Deliver the steak dish in hand to [delivery]''
    \item ``Deliver the steak onion dish in hand to [delivery]''
\end{itemize}

\noindent \textbf{Miscellaneous Actions:}
\begin{itemize}
    \item ``Wait for 5 timesteps''
\end{itemize}

Finally, if communication was disabled between the agents, we removed the message history and did not prompt the agent to send a message to the other agent (i.e., we removed the starred content in the prompt outline).
We provide the full prompt outline for Steakhouse below (the raw prompt at an example timestep during evaluation is in the supplementary material):

\begin{tcolorbox}[colback=blue!5!white,
colframe=blue!50!black,
title=Steakhouse Agent Prompt]

    You are [agent name]. Other agents are: [other agents description].
    
    Environment Details: [environment description]

    \tcbline

    The game has the following dishes: steak dish, steak onion dish. The agents are provided with the current and next order required to make. Ingredients for these dishes are obtained from dispensers. 

    The steak dish requires 2 items: 1 cooked meat (steak) and 1 clean plate.
    The steak onion dish requires 3 items: 1 cooked meat (steak), 1 chopped onion, and 1 clean plate.

    After the dish is complete, it must be delivered to a delivery location.

    \tcbline

    Inventory: [inventory]
    
    Environment Details: [kitchen items]
    
    Location Info: [location info]
    
    Order List: [order list]
    
    Action History: [action history]
    
    $^*$Message History: [message history]

    \tcbline
    
    Your objective is to deliver all the dishes from the order list as quickly as possible. Delivering dishes in the correct order of the order list gives \$100, and out of order gives \$20.
    
    First, choose an action from [action list].
    
    $^*$Then, send a message to the other agent.

\end{tcolorbox}
\noindent\begin{minipage}{\textwidth}
\label{fig:agent_state_prompt}
\end{minipage}

\subsection{Prompts for Mutator LLM}
\label{app:mutator_prompt}

The mutator LLM takes a personality prompt and a mutation direction as input and outputs a new personality prompt to bias the behavior towards the given direction.
The prompts for \planqd{} (Random) and \planqd{} (Directional) are located below, respectively.

\begin{tcolorbox}[colback=orange!5!white,colframe=orange!40!black,title=\planqd{} (Random) Mutation Prompt]

[domain knowledge]

\tcbline

The agent currently has the following personality:

[prompt]

Transform the personality to force the agent to play the game optimally with a random strategy. Ensure the new personality is in second person. Keep the new personality brief and to the point. Only return the transformed personality.

\label{fig:qdllm_prompt_random}
\end{tcolorbox}

\begin{tcolorbox}[colback=lime!5!white,colframe=lime!40!black,title=\planqd{} (Directional) Mutation Prompt]

[domain knowledge]

\tcbline

The agent currently has the following personality:

[prompt]

Transform the personality to force the agent to play the game optimally with [mutation direction]. [mutation context]. Ensure the new personality is in second person. Keep the new personality brief and to the point. Only return the transformed personality.

\label{fig:qdllm_prompt_directional}
\end{tcolorbox}

The LMX mutator takes two personality prompts as input and creates a new prompt as output.

\begin{tcolorbox}[colback=yellow!5!white,colframe=yellow!40!black,title=\planqd{} (LMX) Mutation Prompt]

[domain knowledge]

\tcbline

Based on the following personalities:

[prompt]

Create a new personality similar to all the ones above. Ensure the new personality is in second person. Keep the new personality brief and to the point. Only return the transformed personality.

\label{fig:qdllm_prompt_lmx}
\end{tcolorbox}

The initial prompt is domain-independent and acts as the first stepping stone for all \planqd{} variations and the Random Batch baseline.
This prompt encourages the agent to perform optimally in the environment without worrying about coordination or communication.
We designed this initial prompt to minimize the behavioral bias provided to the agents initially while still encouraging them to perform as best as possible in the environment. We include the full prompt below.

\begin{tcolorbox}[colback=purple!5!white,colframe=purple!75!black,title=Initial Prompt]

You are always focused on the objective. Your goals are set in stone. You don't talk, coordinate, or listen much to others. Only when absolutely necessary do you communicate. Your plan is simple, and nothing will stop you.

\label{fig:initial_prompt}
\end{tcolorbox}

The mutator prompt uses the same domain knowledge given to the agents to inform its mutation. The mutation direction for \planqd{} (Directional) is a randomly selected vector direction in the measure space that is converted to a language direction. We also provide the mutator with additional context on how to modify the input personality prompt to best achieve the mutation direction.
For example, if our archive has two dimensions, specialization and percent contribution (\sref{sec:measures}), then the vector direction $[-1,1]$ would translate to the mutation direction ``decrease specialization and increase percent contribution''. In the case of percent contribution, the additional context for the mutator would specify how to increase the measure, i.e., by encouraging the agent to work more collaboratively with other agents.

The same format applies for the workload measures. For example, if we want to increase the measure ``Difference in Number of Meat Picked'', we mutate the first agent to ``increase number of meat picked'', and provide additional context for the mutator to encourage the agent to focus on picking up raw meats from the meat dispenser.
All nine of our workload measures had a similar mutation context.
The mutator prompt applies this modification to the input personality prompt and generates a new personality prompt as output.

\subsection{Low-level Planner}
\label{sec:low_level_planner}

Once our LLM-based agent determines the high-level action it will take (i.e., pick up raw meat), the selection action is converted to a positional goal that contains the location that the agent needs to move to (in this case, the meat dispenser).
The planner then converts this goal to a set of actions that the agent must take to satisfy this goal.
If all agents haven't moved in the previous timestep, the planner detects it as a collision and resolves it by selecting a random \textit{movement} action (North, South, East, West, or Stay) until the agents can move toward their goal without a collision.

\section{User Studies}

\subsection{User Study 1 (Teaming Data Collection)}
\label{app:user_study1}

We conducted a user study to examine how communication and spatial layout influence human collaboration in a cooperative cooking task. Each session involved a pair of participants who played four rounds across distinct kitchen layouts (open, ring, hallway, and forced coordination).
Participants played the game using a web-based interface, with movement and actions controlled via keyboard inputs. 
In the ``with communication'' condition, participants could verbally coordinate over Zoom, while in the ``without communication'' condition, players had no means of communication and had to rely solely on in-game interactions to collaborate.

\noindent\textbf{Study Design:} We conducted a $2 \times 1$ between-subjects user study to investigate how humans collaborate with and without communication. Each participant played four rounds via a web-based interface with keyboard controls across four kitchen layouts (mentioned earlier) to assess the impact of spatial constraints on coordination. Participants were randomly assigned to either (1) ``with communication'', where verbal interaction was allowed over Zoom, or (2) ``without communication'', where players relied solely on in-game interactions.

\noindent\textbf{Study Protocol:} Participants were recruited from the university campus and were compensated \$10 for their participation.
Each session involved a pair of users, consisting of three phases: onboarding, main study, and exit phase.

In the onboarding phase, participants joined a Zoom session, provided informed consent, and separately completed a pre-experiment survey. They were then introduced to the game rules via a weblink and played a tutorial round on a smaller map in a single-agent setting (i.e., playing alone) to familiarize themselves with the controls. Participants could repeat the tutorial as many times as needed until they felt comfortable with the game dynamics. In the main study phase, participants played four rounds, with layout order and experimental conditions fully counterbalanced to minimize ordering and learning effects. After each round, they completed a brief post-round survey individually. In the exit phase, participants completed a final post-experiment survey, followed by a brief exit interview.

\noindent\textbf{Data Collection:} We had 54 participants (28 males, 26 females) with 27 pairs of teams in our IRB-approved study, ranging in age from $19$ to $38$ $(M = 23.8, SD = 3.6)$. Two participants were excluded from the final analysis due to technical issues that prevented them from completing the fourth round. Among the remaining data, 14 teams participated in the communication condition and 12 in the no-communication condition. To ensure balanced comparison, we included data from 48 participants (24 with communication, 24 without) in our analysis. Our remaining participants comprised of 23 females and 25 males, and our age range remained from $19$ to $38$ ($M = 23.8, SD = 3.8$). Layouts and conditions were fully counterbalanced to mitigate ordering and learning effects—i.e., both layout and condition assignments were randomized and evenly distributed.

\subsection{User Study 2 (Human-likeness Evaluation)}
\label{app:user_study2}

We conducted a second user study after our \planqd{} experiments to verify whether the behavior (without communication) of teams generated by \planqd{} was human-like.

\noindent\textbf{Study Design:}
We ran a within-subjects user study, where participants had to determine the human-likeness of three types of agents based on their gameplay videos. 

The first set of videos was from previously collected human data (User Study 1).
The second set of videos was gameplay from LLM-powered agents generated by \planqd{}.
Finally, the third set of videos was from letting heuristic agents play the game.
Heuristic agents were AI agents with the same action hierarchy as LLM-powered agents, but instead of querying an LLM for high-level actions, they were pre-programmed with rules to follow to select appropriate high-level actions at different states.
The human and heuristic videos acted as grounding for human-like and non-human-like behavior, respectively.

\noindent\textbf{Study Protocol:}
To evaluate whether \planqd{} agents exhibit human-like behavior in our collaborative cooking domain, we conducted an online study using the Prolific platform (prolific.co).
We pre-screened for high-quality participants who had at least 100 approved submissions on Prolific with a 95\% or higher approval rate.

The online study started with onboarding, during which users were shown a human video and a heuristic video, and a brief description of the Steakhouse environment (including \fref{fig:domain_details}). 
These onboarding videos served as ground truth examples to help participants calibrate their understanding of human-like behavior in the task.

Following onboarding, participants completed the main task, where they were shown 12 videos (4 human, 4 \planqd{}, and 4 heuristic) in a randomized order.
After each video, users were asked three questions:

\begin{itemize}
\item \textbf{Q1.} Was this team human or AI? 
\textit{Response:} Binary choice: ``Human'' or ``AI''.
\item \textbf{Q2.} How human-like was this behavior?  
\textit{Response:} 7-point Likert scale (1 = Extremely AI-like, 7 = Extremely human-like).
\item \textbf{Q3.} An attention check question based on the video content.
\end{itemize}

At the end of the video tasks, participants provided their demographic information and an open-ended response about how they classified the behavior. The study took approximately 12 minutes to complete, and the participants were compensated \$1.9 plus a bonus of up to \$0.3 depending on the percentage of attention checks they answered correctly.

\noindent\textbf{Data Collection:}
A total of 40 participants completed the second user study (22 male, 18 female) with ages ranging from 19 to 89 ($M=41.0, SD=13.1$).
Seven participants were excluded from our analysis, six due to a low number of correct answers on the attention check questions, and one due to an incomplete response, resulting in 33 responses.

\section{Experiment Details}

\subsection{Measurements}
\label{app:measures}

We provide additional details about measures from \sref{sec:measures} below:

\subsubsection{Subjective Teaming Measures} 

When collecting human teaming data (User Study 1), we administered post-round and post-experiment questionnaires consisting of 7-point Likert-scale items derived from established measures in human-agent interaction and team collaboration~\citep{mcallister1995affect, hart1988development, hoffman2019evaluating, ryan2000self}. 
From these, we identified the following subjective measures of the team: \textit{Trust, Fluency, Coordination, Satisfaction, and Workload}.  

To identify these underlying constructs, we conducted a principal component analysis (PCA) on responses across all items. We retained components with eigenvalues greater than 1 using the Kaiser criterion and applied varimax rotation to improve interpretability. Items were grouped into scales if they loaded with a correlation of $r \geq 0.6$ on the same factor~\citep{hoffman2020primer}. The resulting scales and their reliability scores are summarized in Table~\ref{table:scales_table}.

\begin{table}
    \centering
    \begin{tabular}{p{0.95\linewidth}}
    \hline
    \textbf{Team Trust} (Cronbach's $\alpha$ = 0.96) \\
    1. My teammate was trustworthy. \\
    2. My teammate’s actions were reliable and predictable. \\
    3. My teammate was committed to the task. \\
    4. I felt confident in my teammate's abilities. \\
    5. I felt comfortable depending on my teammate. \\
    6. I felt synchronized with my teammate’s actions. \\
    \hline
    \textbf{Team Fluency} (Cronbach's $\alpha$ = 0.75) \\
    1. The team worked fluently together. \\
    2. The team’s fluency improved over time. \\
    3. The collaboration contributed to the fluency and better performance of the interaction. \\
    4. The interaction felt natural and effortless. \\
    \hline
    \textbf{Coordination} (Cronbach's $\alpha$ = 0.85) \\
    1. I had to carry the weight to make the team better. \\
    2. I was the most important member of the team. \\
    \hline
    \textbf{Satisfaction} (Cronbach's $\alpha$ = 0.73) \\
    1. I enjoyed the gameplay experience. \\
    2. The task was engaging and immersive. \\
    \hline
    \textbf{Demand} (Cronbach's $\alpha$ = 0.76) \\
    1. How mentally demanding was the game? \\
    2. How hurried or rushed was the pace of the game? \\
    \hline
    \textbf{Likert items are coded as 1 (Strongly Disagree) to 7 (Strongly Agree)} \\
    \end{tabular}
    \captionsetup{justification=centering}
    \caption{{\small Subjective Scale Measure Items.}}
    \label{table:scales_table}
\end{table}

\subsubsection{Teamwork Measures} 
Below are calculations for each teamwork measure:
\begin{itemize}
    \item \textbf{Percent contribution:} This is calculated by the formula $\frac{1}{n} \sum^n_i D_i$, where $D_i$ measures the amount a team worked together on a particular delivered dish, and $n$ is the total number of dishes delivered. The final contribution value for a specific delivered dish ($D_i$) is calculated using the formula:

    $$\frac{\min{(n_{P_1},n_{P_2})}}{n_{P_1}+n_{P_2}}$$
    
    where $n_{P_j}$ is the number of high-level actions for a specific delivered dish taken by player $j$. Across a single game, we calculate this result for each completed dish, and take the average over all results as the final value. This measure ranges from $[0,0.5]$. A score of 0 means that a dish was completed by only one player, and higher scores means the work for the dish was more distributed across both players.

    \item \textbf{Specialization:} We first define four \textit{action groups} $A_1, \dots, A_4$: ingredient actions, plate actions, dish creation actions, and delivery actions. 
    All high-level actions are bundled into one of the following action groups. 
    For example, picking up a raw meat or onion falls into the ingredient action group, whereas creating a steak or steak with onion dish falls into the dish creation group. 
    The specialization measure is obtained from the formula $\frac{1}{N} \sum_i^N Sp_i$, where $Sp_i$ measures how much a player chose actions from a specific action group, and $N$ is the number of players.
    $Sp_i$ is calculated using the formula:

    $$\frac{\max{(n_{A_1},...,n_{A_4})}}{\sum^4_{k=1} n_{A_k}}$$

    where $n_{A_k}$ is the number of times a player has taken an action from the action group $A_k$. This value has a range of $[0.25, 1]$. A specialization value of 0.25 means that each player evenly distributed their actions across all action groups, and an increasing specialization value means that each player primarily only did one type of action in the action group.

    \item \textbf{Fitness:} We use the discounted sum of rewards $J = \sum_t \gamma^t r_t$ defined in \sref{sec:problem}, where $r_t$ is the reward defined in \apref{app:domain}. For our experiments, we used $\gamma=1$.

    \item \textbf{Average action delay:}
    Action delay is defined as the number of timesteps between non-movement actions (interactions) conducted by \textit{either} agent. For example, this means that if one agent does not perform any interaction actions during the episode, the action delays are only measured from the second agent. Average action delay is the mean action delay across the whole episode.
\end{itemize}

\subsection{Algorithms} 
\label{app:algos}

We compare all \planqd{} variations and Random Batch in our experiments.
All experiments were given a budget of 100 prompt evaluations, with each evaluation being repeated four times and aggregated with the median.

\subsubsection{PLAN-QD}

We divided the budget of 100 prompt evaluations into 50 iterations of the QD loop.
The batch size was 2, i.e., at each iteration, we selected and mutated 2 prompt lists (i.e., 4 individual personality prompts) to be evaluated in our domain.

In our Steakhouse simulation, the re-query timeout was 5 timesteps, meaning agents chose a new high-level action after 5 timesteps of being idle. This re-query happened if the agents' action became invalid for 5 timesteps or if they did not perform any action for 5 timesteps.
Furthermore, the action and message history parameters were both set to 2. 
The message history parameter is the number of past messages by either agent that an agent can see in the prompt.
The action history parameter is the number of \textit{completed} actions by the agent that is in the prompt.
If the agent selected a high-level action but never completed it, it was not provided in their action history.
We also provided an associated relative timestep to the actions and messages to give the agents more context.
For example, if a message is sent at timestep $t$, and received at timestep $t+k$, we mentioned in the prompt that the message was sent by an agent $k$ timesteps ago.

\subsubsection{Random Batch}

The Random Batch baseline directly queried the mutator LLM for 100 prompt lists (i.e., 200 individual personality prompts) with the following mutation prompt:

\begin{tcolorbox}[colback=brown!5!white,colframe=brown!75!black,title=Random Batch Prompt]

[domain knowledge]

\tcbline

The agent currently has the following personality:

[initial prompt]

Create [batch size * 2] random personalities for the agent to play the game optimally with a random strategy. Ensure the new personality is in second person. Keep the new personalities brief and to the point. 

\label{fig:random_mutation_prompt}
\end{tcolorbox}

Domain knowledge and the initial prompt are exactly the same as in \planqd{}.

\begin{table*}
    \centering
    \footnotesize

    \begin{tabular}{lcccccccc}
        \toprule
        & \multicolumn{2}{c}{Open} & \multicolumn{2}{c}{Ring}
        & \multicolumn{2}{c}{Hallway} & \multicolumn{2}{c}{Forced Coordination} \\
        \cmidrule(lr){2-3} \cmidrule(lr){4-5} \cmidrule(lr){6-7} \cmidrule(lr){8-9}
        & w/o Comm. & w/ Comm. & w/o Comm. & w/ Comm. & w/o Comm. & w/ Comm. & w/o Comm. & w/ Comm. \\
        \midrule
        
        \textbf{Random Batch} & $10.83_{3.41}$ & $12.90_{3.75}$ & $8.26_{1.21}$ & $8.54_{1.30}$ & $8.41_{1.46}$ & $7.91_{1.23}$ & $1.91_{0.90}$ & $1.94_{1.06}$ \\

        \textbf{PLAN-QD} & & & & & & & & \\

        \textbf{\quad(Random)} & $11.41_{3.38}$ & $12.25_{4.11}$ & $7.89_{1.09}$ & $8.56_{1.30}$ & $7.54_{1.27}$ & $7.41_{1.34}$ & $1.82_{1.05}$ & $1.90_{0.94}$ \\
        
        \textbf{\quad(LMX)} & $12.83_{3.53}$ & $12.07_{3.28}$ & $7.21_{0.72}$ & $8.11_{1.04}$ & $8.12_{1.03}$ & $7.59_{1.36}$ & $1.69_{0.88}$ & $1.64_{1.03}$ \\

        \textbf{\quad(Directional)} & $\textbf{13.39}_{4.50}$ & $\textbf{15.07}_{4.73}$ & $\textbf{9.12}_{1.54}$ & $\textbf{10.27}_{1.67}$ & $\textbf{11.29}_{2.58}$ & $\textbf{11.78}_{2.49}$ & $\textbf{2.58}_{1.14}$ & $\textbf{3.08}_{1.19}$ \\

        \bottomrule
    \end{tabular}

    \caption{Average QD-scores across distinct pairs of measures. \planqd{} with directional mutation obtains the highest QD-score in all layouts and communication conditions. Values are in terms of thousands.}
    \label{tab:qd_versus_random_qd_score}
\end{table*}

\subsection{LLM Setup}

For our primary experiments, both our mutator LLM and the LLM powering the agents were Meta's LLaMA 3.1 70B Instruct~\citep{grattafiori2024llama} model hosted in the SGLang framework ~\citep{sglang}. 
We include model ablations for the mutator and the agent in \sref{ssec:qdllm_benchmark} and \apref{app:agent_llm_ablations}, respectively.
For each algorithm, all conditions (four layouts and two communication conditions) were run in parallel.
In each experiment, the four repetitions of two evaluations were run in parallel, resulting in a total of 64 parallel episodes in the Steakhouse environment across all parallel experiments.
Each episode required 150 to 250 total LLM queries in sequence.
Our experiments utilized 8 A6000 GPUs, and took between 6-9 days depending on the layout. Forced coordination experiments took the longest and hallway experiments took the shortest amount of time.

The LLM queries used a temperature of 1.1 and the default parameters provided in the SGLang framework, which include \texttt{top\_k = -1}, \texttt{top\_p = 1.0}, and \texttt{max\_new\_tokens = 128}.

\section{Additional Results}

\subsection{Effects of Layout and Personality on Human Teaming}
\label{app:user_study_res}

In addition to the analysis presented in \sref{ssec:results_user_study}, we tested the effect of layout and personality on the teamwork measures in the collected human teaming data (User Study 1).

\subsubsection{Effect of Layout}

Beyond fitness, we examined how layout influenced additional coordination and efficiency metrics. 
We conducted a two-way ANOVA with layout (4 levels: open, ring, hallway, forced coordination) and communication condition (2 levels: with communication, without communication) as between-subject factors.
The results indicate that layout significantly shaped team coordination strategies.
A significant main effect of layout was observed on percentage contribution, $F(3,184)=8.41,p<.001$, suggesting that spatial constraints influenced how much each player contributed to task completion.

Participants in the forced coordination layout contributed significantly more $(M=0.36)$ than those in other layouts, while the ring layout resulted in the lowest contributions $(M=0.23)$. Communication did not significantly impact contribution levels, $F(1,184)=0.63,p=0.43$.

The effect of layout on task specialization was also highly significant, $F(3,184)=7.36,p<.001$. Participants exhibited the highest degree of specialization in the forced coordination layout $(M=0.81)$, while open and ring layouts led to significantly lower specialization $(M=0.62)$ and $(M=0.61)$, respectively. This suggests that constrained layouts encouraged division of labor, whereas open spaces allowed for more fluid role-switching.

Furthermore, we evaluate the effect of total movement actions in all layouts, which is the total number of times a player moved in an episode. Layout had a highly significant effect on movement actions, $F(3,184)=22.04,p<.001$. The forced coordination layout resulted in the least movement actions $(M=112)$, while hallway layouts led to the most movement $(M=182)$, indicating that spatial constraints influenced the efficiency of player movements. However, communication did not significantly impact movement, $F(1,184)=1.45,p=0.232$, nor was there a significant interaction between layout and communication.

Overall, spatial constraints play a significant role in shaping coordination strategies and team efficiency. Constrained environments, such as the forced coordination layout, encouraged greater individual contributions and role specialization but also led to inefficiencies, such as increased wasted actions. In contrast, open layouts supported more flexible role-switching and dynamic task-sharing. These findings highlight that the structure of the environment can strongly influence human teaming behavior—often outweighing the effects of communication. Therefore, future studies aiming to isolate the effects of communication on teamwork should carefully account for spatial constraints when designing experimental setups.

\subsubsection{Effect of Personality}
\label{app:personality_differences}
Beyond spatial constraints of layouts, player differences such as background knowledge and different personalities also influenced communication effects. Participants with stronger game knowledge (evidenced by higher tutorial performance) exhibited a communication effect (likely due to a teaching effect) in the communication condition. This leads to a $20.7\%$ decrease in the average action delay in the forced coordination layout ($t(12) = -1.20$, $p=0.24$). Additionally, in the no-communication condition, teams showed a strong Pearson correlation between score and Trust scale in the ring ($r(12) = 0.65$) and hallway ($r(12) = 0.71$) layouts, suggesting that greater trust facilitated implicit coordination across shared counters, leading to improved performance.

\subsection{QD-scores Obtained by PLAN-QD and Baselines}
\label{app:coverage_res}

We provide the averaged QD-scores (\sref{sec:problem}) across the $C(12,2)=66$ distinct pairs of measures in \tref{tab:qd_versus_random_qd_score}. These results highlight that the additional behaviors covered by \planqd{} with directional mutation, compared to other baselines, are also high-performing. The CSV files with the full table containing the unaggregated coverage and QD-score values are in the supplementary material.

\subsection{Agent Model Ablations}
\label{app:agent_llm_ablations}

\begin{table}
    \centering
    \footnotesize
    \begin{tabular}{lcc}
        \toprule
        \textbf{Model} & \textbf{w/ Comm.} & \textbf{w/o Comm.} \\
        \midrule
        \multicolumn{3}{c}{\textbf{Open}} \\
        \midrule
        LLaMA 3.1 8B Instruct & $0_{0.0}$ & $0_{0.0}$  \\
        LLaMA 3.1 70B Instruct & $425_{65.0}$ & $305_{35.6}$  \\
        Gemini 2.5 Flash & $\textbf{475}_{21.7}$ & $\textbf{475}_{21.7}$ \\
        \midrule
        \multicolumn{3}{c}{\textbf{Forced Coordination}} \\
        \midrule
        LLaMA 3.1 8B Instruct & $25_{21.7}$ & $0_{0.0}$  \\
        LLaMA 3.1 70B Instruct & $\textbf{450}_{25.0}$ & $\textbf{425}_{41.5}$  \\
        Gemini 2.5 Flash & $165_{27.7}$ & $155_{42.1}$ \\
        \bottomrule
    \end{tabular}
    
    \caption{Mean fitness for different agent LLMs across 4 runs. Standard error is provided in subscripts.
    LLM capability affects the peak performance of the corresponding teams.}
    \label{tab:agent_model_ablations_fitness}
\end{table}

To study the effect of various LLMs of team performance, we fixed the personality prompts used to seed \planqd{} and varied the agent LLM.
\tref{tab:agent_model_ablations_fitness} shows the average team performance (fitness) across two layouts. We observed that LLaMA 8B, a smaller model than other baselines, was unable to deliver dishes in our collaborative environment, suggesting it lacked the planning or reasoning capacity needed for long-horizon coordination.
Interestingly, performance also varied across larger LLMs, suggesting that average team fitness depends not only on model size but also on the specific strengths and capabilities of the underlying LLM.

\clearpage

\end{document}